\documentclass[conference]{IEEEtran}
\IEEEoverridecommandlockouts
\usepackage{cite}
\usepackage{amsmath,amssymb,amsfonts}
\usepackage{algorithmic}
\usepackage{graphicx}
\usepackage{textcomp}
\usepackage{xcolor}
\usepackage{hyperref}
\usepackage{url}
\usepackage{subcaption}

\def\BibTeX{{\rm B\kern-.05em{\sc i\kern-.025em b}\kern-.08em
    T\kern-.1667em\lower.7ex\hbox{E}\kern-.125emX}}
\begin{document}

\title{PokeRL: Reinforcement Learning for Pokémon  Red\\
}

\author{\IEEEauthorblockN{Dheeraj Reddy Mudireddy}
\IEEEauthorblockA{\textit{Texas A\&M Institute of Data Science} \\
\textit{Texas A\&M University}\\
College Station, USA \\
dheeraj.reddy@tamu.edu}
\and
\IEEEauthorblockN{Sai Patibandla}
\IEEEauthorblockA{\textit{Dept. of Computer Science} \\
\textit{Texas A\&M University}\\
College Station, USA \\
patibas@tamu.edu}
}

\maketitle

\begin{abstract}
Pokémon Red is a long-horizon JRPG with sparse rewards, partial observability, and quirky control mechanics that make it a challenging benchmark for reinforcement learning. While recent work has shown that PPO agents can clear the first two gyms using heavy reward shaping and engineered observations, training remains brittle in practice, with agents often degenerating into action loops, menu spam, or unproductive wandering. In this paper we present PokeRL, a modular system that trains deep reinforcement learning agents to complete early-game tasks in Pokémon Red, including exiting the player’s house, exploring Pallet Town to reach tall grass, and winning the first rival battle. Our main contributions are a loop-aware environment wrapper around the PyBoy emulator with map masking, a multi-layer anti-loop and anti-spam mechanism, and a dense hierarchical reward design. We argue that practical systems like PokeRL, which explicitly model failure modes such as loops and spam, are a necessary intermediate step between toy benchmarks and full “Pokémon League champion” agents. Code is available at \url{https://github.com/reddheeraj/PokemonRL}.
\end{abstract}

% \begin{IEEEkeywords}
% component, formatting, style, styling, insert
% \end{IEEEkeywords}

\section{Introduction}
Classic game environments have been central to the progress of deep reinforcement learning, from Atari and Go to NetHack and Minecraft. These domains combine high-dimensional observations with long time horizons and have repeatedly revealed the limitations of standard RL algorithms in the presence of sparse rewards and deceptive local optima. 

Pokémon Red occupies a particularly interesting point in this landscape. Unlike reflex-oriented arcade games, it is a story-driven JRPG where progress depends on coordinated navigation, interaction, and turn-based combat, executed over tens of thousands of timesteps. The player must leave their house, traverse Pallet Town and nearby routes, collect a starter Pokémon, engage in wild battles to gain experience, and defeat scripted opponents. From an RL perspective, this yields:

\begin{itemize}
    \item \textbf{Extreme rewards assignment}: Key rewards such as catching a Pokémon or winning a battle may occur only after long sequences of actions.

    \item \textbf{High branching factor}: Even in the opening minutes there are many movement options and interaction opportunities.
    
    \item \textbf{Partial observability}: The agent sees only the current 2D screen, while crucial variables (HP, map ID, party status) live in hidden memory.

    \item \textbf{Non-standard controls}: Movement requires a double press (turn then step), and multiple menu buttons exist that are useless or harmful for learning.
\end{itemize}

Recent work has shown that PPO agents can complete roughly the first fifth of Pokémon Red by using a custom Gym environment, dense reward shaping, and an additional “visited mask” observation channel \cite{Rubinstein2025}. However, reproducing these results is non-trivial. In our own attempts, we observed several recurring pathologies: agents exploiting shaping rewards by looping, spamming the A button or the no-op action, spinning in place because the environment mishandled double-press movement, and wandering within a small region without meaningful exploration.

PokeRL is our attempt to build a more robust and transparent environment for early-game Pokémon Red. Instead of only tuning rewards, we explicitly encode loop detection, spam penalties, and spatial memory into the environment. We then structure training as a curriculum over three sequences: exiting the house, reaching tall grass from Pallet Town, and winning the first rival battle. The result is not a full game-solving agent, but a working system that reliably acquires non-trivial behaviors and surfaces many of the design trade-offs involved in turning “RL + emulator” into a practical research platform.

\section{Motivation and Problem Setting}

\subsection{Why Pokemon Red?}

\begin{figure*}[t]
    \centering
    \begin{subfigure}[b]{0.32\linewidth}
        \centering
        \includegraphics[width=\linewidth]{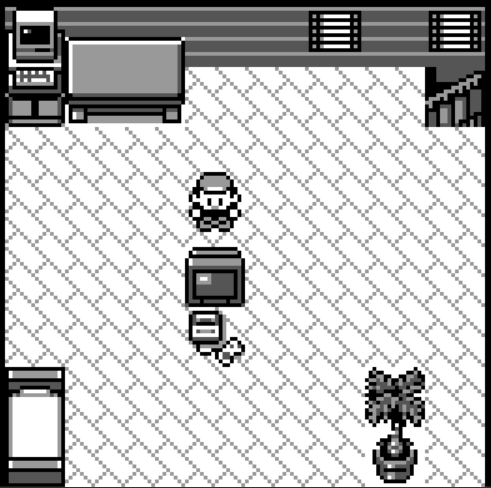}
        \caption{Red in his house}
        \label{fig:red_in_house}
    \end{subfigure}
    \hfill
    \begin{subfigure}[b]{0.32\linewidth}
        \centering
        \includegraphics[width=\linewidth]{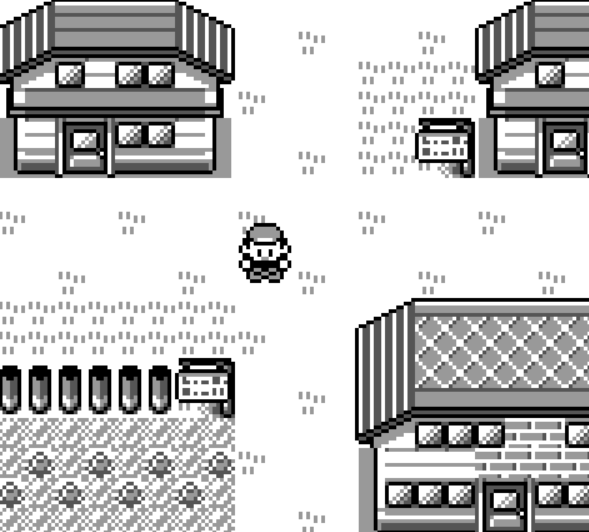}
        \caption{Exploring Pallet Town}
        \label{fig:pallet_town}
    \end{subfigure}
    \hfill
    \begin{subfigure}[b]{0.32\linewidth}
        \centering
        \includegraphics[width=\linewidth]{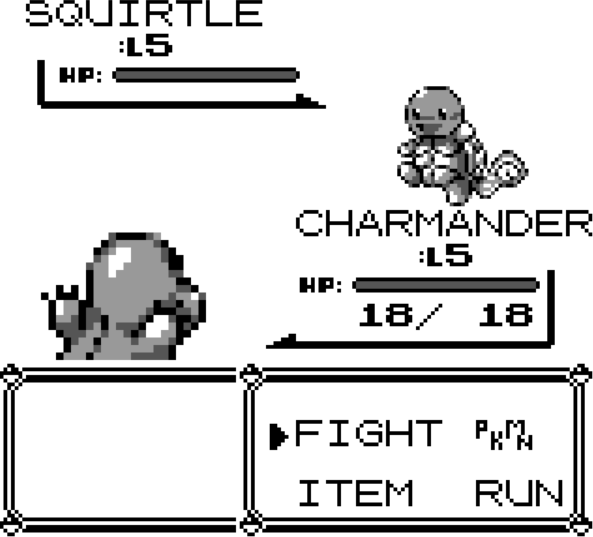}
        \caption{First Battle}
        \label{fig:red_vs_blue}
    \end{subfigure}
    \caption{Frames of selected objectives in the game.}
    \label{fig:sequences}
\end{figure*}

Pokémon Red is a natural testbed for long-horizon RL because it embeds several sub-tasks commonly studied in isolation:

\begin{itemize}
    \item Navigation and exploration in a 2D grid-like overworld.
    \item Inventory and resource management through items and Pokémon party composition.
    \item Turn-based combat with discrete actions, stochastic outcomes, and non-trivial strategy as shown in Figure \ref{fig:red_vs_blue}.
\end{itemize}

% \begin{figure}
%     \centering
%     \includegraphics[width=0.7\linewidth]{Red vs Blue.png}
%     \caption{Charmander vs Squirtle, in the initial sequence where main character fights his rival.}
%     \label{fig:red_vs_blue}
% \end{figure}

The opening segment of the game provides a compact yet challenging subset of these skills:

\begin{enumerate}
    \item \textbf{House exit}: Learn basic movement and door interaction to leave Red’s bedroom and house.

    \item \textbf{Early exploration}: Navigate Pallet Town and Route 1 to reach tall grass and trigger scripted encounters.

    \item \textbf{First rival battle}: Execute viable battle actions to defeat the rival’s Pokémon given a fixed starter.
\end{enumerate}

Our project focuses on these three tasks because they are short enough to allow experimentation on a single device and rich enough to expose the core issues of loop, sparse rewards, and bad exploration.

\subsection{Challenges in Vanilla RL Setup}
Preliminary experiments with a simple PPO agent wrapped around PyBoy highlighted five concrete failure modes:

\begin{itemize}
    \item \textbf{Infinite action loops}: The agent discovered local cycles (for example, pacing between two tiles) that gave small positive movement rewards without progress.

    \item \textbf{Reward sparsity and instability}: When only rare events were rewarded, episode returns fluctuated wildly, and policies failed to improve.

    \item \textbf{Button spam}: Action distributions collapsed onto A, Start, or no-op, producing pointless menus or idling.

    \item \textbf{Incorrect movement semantics}: Treating a single direction press as “move one tile” caused the agent to spin in place instead of walking.

    \item \textbf{Exploration without memory}: Without an explicit notion of visited states the agent repeatedly revisited already explored areas.
\end{itemize}

PokeRL is designed as an answer to these five issues.

\section{Related Work}
Deep RL in video games began with agents that achieved human-level performance on Atari games by learning directly from pixels with score-based rewards. Games such as Montezuma’s Revenge exposed the limitations of this approach: rewards were extremely sparse, and naive exploration strategies failed to reach them. Work on intrinsic motivation and Go-Explore \cite{goexplore2021} addressed this by explicitly rewarding novelty and maintaining state memories to revisit promising situations.
\\

Until recently, little academic literature existed on applying RL to Pokémon. For Pokémon specifically, Pleines et al \cite{pleines2025} introduced the first comprehensive deep RL agent for Pokémon Red, using PPO \cite{schulman2017} over a PyBoy-based Gym environment. They engineered a reward function with more than 25 components and augmented the observation with a 72×80 visited mask channel, enabling an agent to progress up to the second gym in Cerulean City, \textbf{beating two major gym leaders in pokemon battles}. This was a milestone achievement: as of February 2025, an RL policy (with $<$10 million parameters) could beat Pokémon Red’s early game to an extent with only a few simplifications to the game logic. They showed that certain reward bonuses could backfire, as agents learned to exploit them in unintended ways. Their ablation studies showed both the necessity and risk of dense shaping: certain rewards encouraged grinding or skipping mandatory events when tuned poorly.

Parallel community projects have applied RL to other Pokémon titles, such as Pokémon Gold, and constructed tools like poke-env \cite{poke_env}, which interfaces RL algorithms with the Pokémon Showdown battle simulator. These battle-only settings provide cleaner reward signals (win or loss) and smaller action spaces, and have been pushed to near human-parity using offline RL and large language model policies. For example, PokeLLMon (2024) \cite{hu2024pokellmon} integrated a large language model as the decision-making policy and reached human-parity in Pokémon Showdown battles by leveraging prior knowledge of move effectiveness and strategy. These battle-focused studies, while not addressing the exploration aspect, contribute insights on reward design (win/loss or HP-based rewards) and action masking (disabling illegal moves) that ensure the RL agent learns valid battle strategies.

More recently, the NeurIPS 2025 PokéAgent Challenge \cite{karten2025pokeagent} proposed tracks for Pokémon Emerald speedrunning and competitive play, positioning Pokémon as a benchmark for sequential decision making.

Adding onto the community projects, in 2023, a YouTube video demonstrating an RL agent playing Pokémon went viral (7.5+ million views) \cite{whidden2023pokemonRL}. That project was built on PyBoy and OpenAI Gym as well, and it inspired many subsequent experiments. Bloggers and engineers have tried related ideas on other Pokémon titles; for instance, Bas de Haan documented a project training an agent on Pokémon Gold version \cite{deHaan2024}. In his write-up, he emphasizes the importance of reading emulator memory for key game variables (to compute custom rewards for things like experience points or map exploration).

Our work is closest in spirit to the open PokeRL framework and its academic counterpart, but focuses on explicitly engineered anti-loop and anti-spam mechanisms, a per-map visited mask with centered origin, and a curriculum over early-game sequences rather than full game completion.

\section{System Overview}
PokeRL is built as a custom Gymnasium environment around PyBoy with additional modules for memory reading, reward computation, and curriculum control. 

\subsection{Actor-Critic Network}
\begin{figure*}
\centerline{\includegraphics[width=0.7\linewidth]{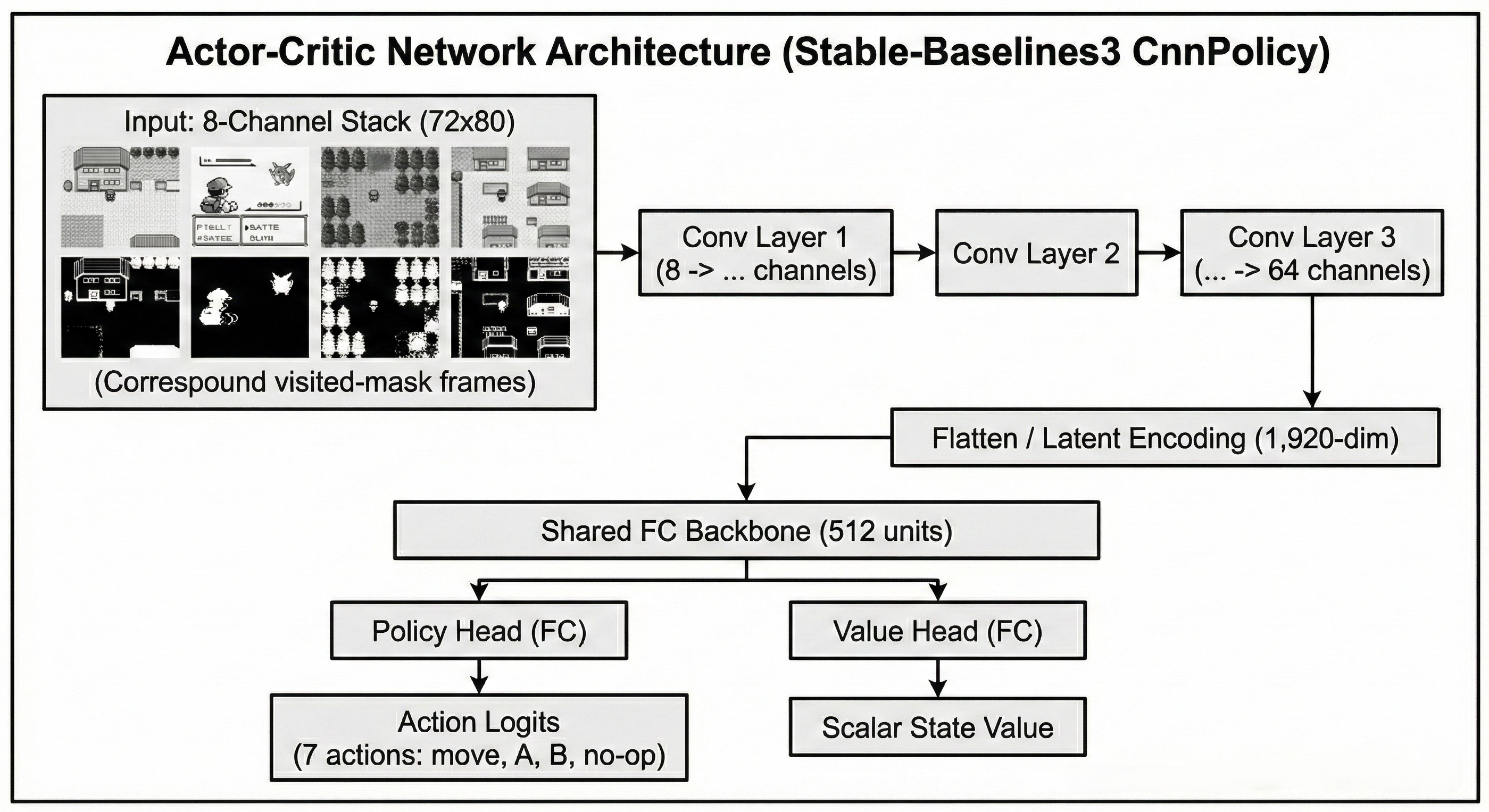}}
\caption{The Actor-Critic Network}
\label{model}
\end{figure*}

The agent’s policy and value functions are implemented using the Stable-Baselines3 CnnPolicy, which provides a compact yet expressive convolutional actor–critic model well suited for visual environments. As displayed in Figure \ref{model}, the network processes an 8-channel input composed of four consecutive grayscale game frames and their corresponding visited-mask frames, to capture both short-term temporal dynamics and spatial exploration history. Three convolutional layers progressively downsample the 72×80 input while expanding feature depth from 8 to 64 channels, producing a 1,920-dimensional latent encoding. This representation is then passed through a fully connected layer with 512 units, forming the shared backbone for both the policy and value heads. The policy head outputs logits over the seven allowed actions (movement, A, B, and no-op), while the value head predicts the scalar state value. The model contains $\approx$ \textit{1.07 million trainable parameters}, making it lightweight enough for fast iteration while still capable of learning structured behaviors.

\subsection{Observation Space}
\begin{figure}
    \centering
    \includegraphics[width=0.8\linewidth]{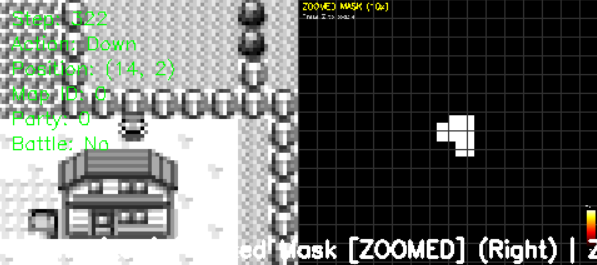}
    \caption{downsampled frame and its visited mask.}
    \label{fig:visited_mask}
\end{figure}

The base observation is a downsampled grayscale image of the Game Boy screen (72×80) as shown in Figure \ref{fig:pallet_town}. To give the agent short-term temporal context and long-term spatial memory, we augment this as follows:

\begin{enumerate}
    \item \textbf{Frame stacking}: Four consecutive frames are concatenated along the channel dimension.
    \item \textbf{Per-map visited mask}: For each map we maintain a 72×80 binary mask that marks tiles the agent has already visited. The mask is centered at the entry position of that map, so as the camera scrolls, the mask remains aligned with global map coordinates. This is illustrated in Figure \ref{fig:visited_mask}.
\end{enumerate}

The final observation has shape 72×80×2 before stacking and 72×80×8 after stacking, which is then transposed to channels-first for a CNN policy.

\subsection{Action Space and Double-Press Handling}
We restrict the agent to seven actions: up, down, left, right, A, B, and no-op. Start and Select are removed because they open a menu and are a prime target for spam. At one point, the agent learned to save the game after coming to the ground floor in his house.

Pokémon Red uses a double-press mechanic where the first directional press rotates the character and only the second moves the player one tile. We implement this directly in the environment: any movement action issues two press-release cycles to PyBoy. This abstraction ensures that a single RL action corresponds to an intuitive “step” in the grid world.

\subsection{Memory Reader}
To define precise rewards and termination conditions, we read selected addresses from the game’s RAM via PyBoy. The RAM addresses are mentioned in Table \ref{tab:memory}. This information is not passed directly to the policy but is used by the environment to compute events such as “entered tall grass” or “battle started.”

\begin{table}[htbp]
\centering
\caption{Direct Memory Reading Addresses}
\begin{tabular}{|c|c|}
\hline
\textbf{Data} & \textbf{Memory Address} \\ \hline
Player Position (Y) & 0xD361 \\ \hline
Player Position (X) & 0xD362 \\ \hline
Map ID & 0xD35E \\ \hline
Party Count & 0xD163 \\ \hline
Battle State & 0xD057 \\ \hline
Pokémon HP & 0xD16C \\ \hline
\end{tabular}
\label{tab:memory}
\end{table}

\section{Reward Design and Curriculum}
\subsection{Hierarchical Rewards}
We design a dense reward function with three conceptual levels: micro, meso, and macro. Representative terms are shown in the Table \ref{tab:rewards}.

To counterbalance these positives, we add small penalties for staying in place, repeated button presses, and detected loops (section \ref{section:loop}). After several iterations, we settled on relatively mild negative values in the range $-0.02$ to $-0.2$ to avoid overwhelming positive rewards.

\begin{table}[h!]
\centering
\caption{Reward Structure for Agent in Environment}
\begin{tabular}{|l|l|}
\hline
\textbf{Reward Type} & \textbf{Reward Description} \\ \hline
\multicolumn{2}{|c|}{\textbf{Micro Rewards (per step)}} \\ \hline
Movement to a new tile & +1.0 \\ 
Euclidean distance moved & +0.2 $\times$ distance \\ 
Visiting a position for the first time & +0.5 \\ \hline
\multicolumn{2}{|c|}{\textbf{Meso Rewards (sub-goals)}} \\ \hline
Transitioning to a new map & +10.0 \\ 
Entering a map for the first time in an episode & +5.0 \\ 
Exploration (large unexplored region reached) & +2.0 \\ \hline
\multicolumn{2}{|c|}{\textbf{Macro Rewards (rare events)}} \\ \hline
Entering tall grass & +20.0 \\ 
Starting a battle & +10.0 \\  
Catching a Pokémon / winning Sequence 3 battle & +50.0 \\ \hline
\end{tabular}
\label{tab:rewards}
\end{table}

\subsection{Curriculum over Three Sequences}
Instead of training a single policy on the full game start, we split the problem into three sequences, each with its own configuration and save state:

\begin{enumerate}
    \item \textbf{Sequence 1: House Exit}
    \begin{itemize}
        \item Start in Red’s upstairs bedroom.
        \item Episode ends when the map ID changes to indicate outdoors or when a step limit is exceeded.
        \item Rewards emphasize movement and the first map transitions.
    \end{itemize}
    
    \item \textbf{Sequence 2: Exploration to Grass}
    \begin{itemize}
        \item Start outside Red’s front door.
        \item Episode ends when the agent reaches tall grass, triggers Professor Oak’s scripted event, or times out.
        \item Rewards emphasize exploration coverage and reaching the grass region.
    \end{itemize}
    
    \item \textbf{Sequence 3: First Rival Battle}
    \begin{itemize}
        \item Start at the beginning of the rival battle with a fixed starter Pokémon.
        \item Rewards focus on using offensive moves, fainting the opponent’s Pokémon, and winning the battle.
    \end{itemize}
\end{enumerate}

This curriculum mirrors the game’s natural progression and allows each skill to be learned under a simpler reward structure.

\section{Loop and Spam Mitigation} \label{section:loop}
\subsection{Anti-Loop System}
To combat infinite action loops we introduce a three-layer anti-loop mechanism:

\begin{enumerate}
    \item \textbf{Position visit penalties}: We maintain a dictionary of counts for each (x, y) position within an episode. When a tile exceeds three visits the agent receives a small negative reward, and a larger penalty beyond five visits.
    \item \textbf{Action pattern detection}: Using a sliding window over 20 recent actions, we check for repetitive patterns such as alternating A and B or fixed two-action cycles. Repeated detection adds a negative reward, while breaking out of such patterns yields a small bonus.
    \item \textbf{Position loop detection}: We track the last several positions and detect when the agent repeatedly returns within a small radius of a previous location. This is treated as a loop and penalized.
\end{enumerate}

\begin{figure}[h]
    \centering
    \includegraphics[width=\linewidth]{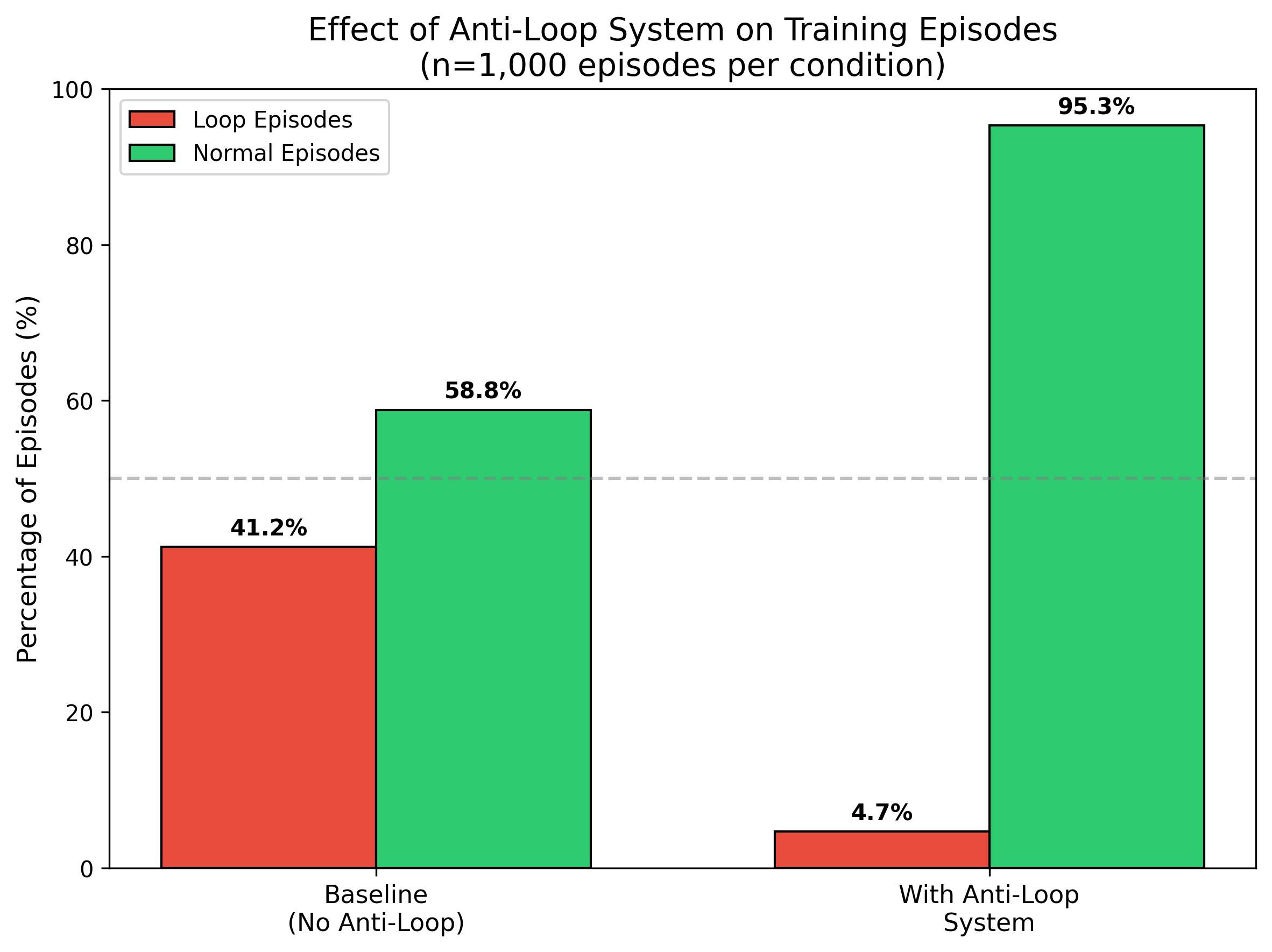}
    \caption{Effect of anti-loop system on training episodes}
    \label{fig:anti-loop}
\end{figure}

This system directly encodes the notion that “doing the same thing in the same place” is unproductive. We instrumented the environment to log the $position\_visits$ and $action\_patterns$ during training. A "\textbf{loop episode}" is any episode where \textbf{a single position was visited $>$10 times OR action pattern detection triggered $>$20 times}. We then analyzed 1000 episodes before and after anti-loop implementation, drawing out the results mentioned in Figure \ref{fig:anti-loop}.

\subsection{Anti-Spam Mechanism}
We observed that PPO policies quickly converged to policies that hammered the A button, no-op, or menu buttons. Naively imposing large penalties for these actions caused episode returns to become highly negative and made training unstable. Our final anti-spam design uses graduated streak penalties:

\begin{itemize}
    \item Streak counters for A, B, and staying in the same position.
    \item Small penalty ($-0.1$) after three consecutive presses of the same button.
    \item Additional penalty ($-0.2$) beyond five presses.
    \item Tiny positive reward when the last several actions include at least four distinct actions, encouraging diversity.
\end{itemize}

We also entirely removed Start and Select from the action space to close the most egregious loophole. 

\begin{figure}
    \centering
    \includegraphics[width=\linewidth]{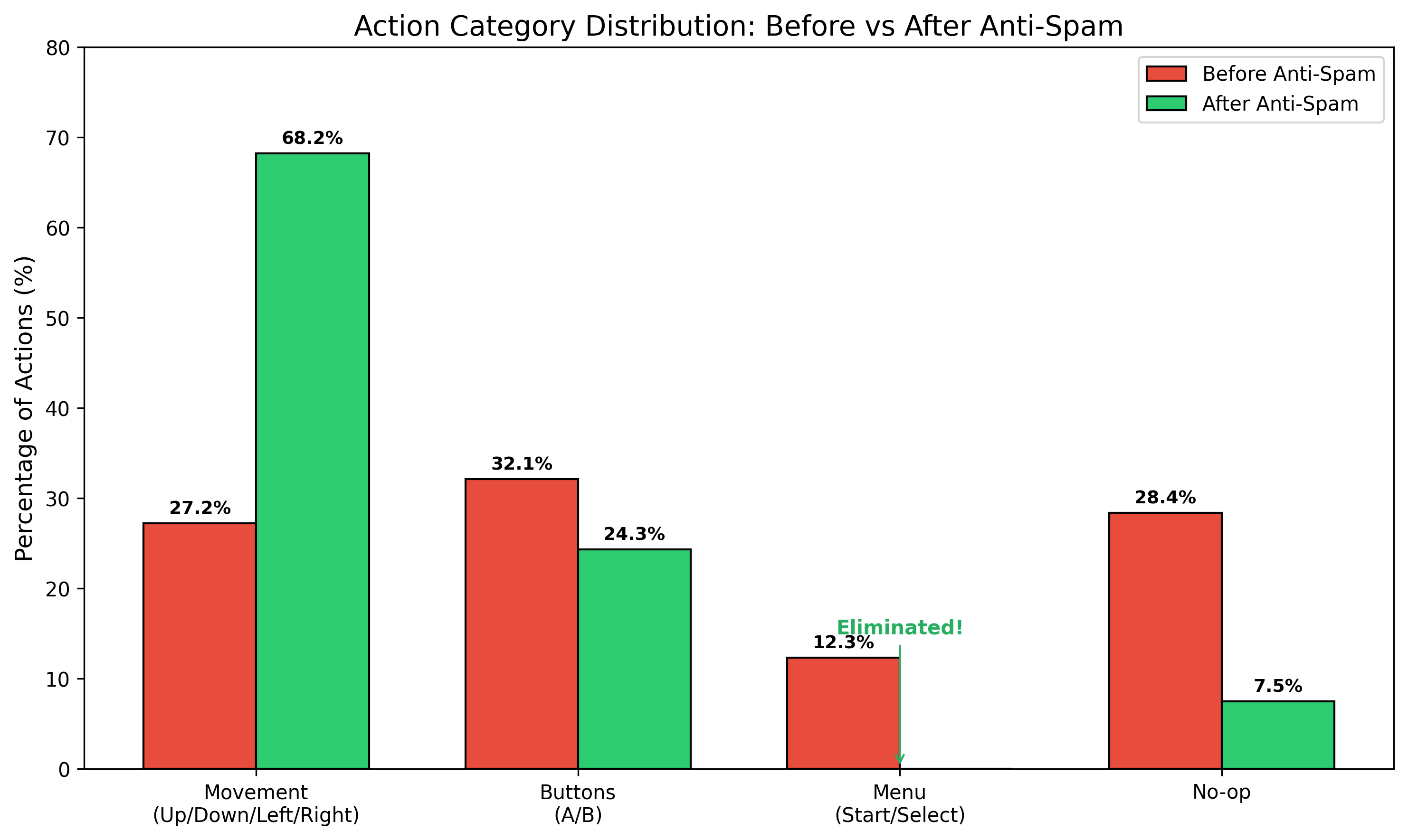}
    \caption{Actions distribution before vs after anti-spam implementation}
    \label{fig:anti-spam}
\end{figure}

To measure the impact of this mechanism, we implemented an $action\_counts$ array in the environment to track frequency of each action. Using this array, we were able to log the action distribution to tensorboard every 5000 steps during training. To statistically distinguish between before vs after anti-loop mechanism, we calculated the Shannon Entropy (H) \cite{SARAIVA2023194}, defined as:

\begin{equation}\label{eq:entropy}
H = - \sum_{a} p(a)\,\log_{2} p(a)
\end{equation}

where $p(a)$ is probability of action $a$. Before anti-spam, $H = 1.21$ bits while after anti-spam implementation, $H = 1.82$ bits. This increase in H indicates that the agent explores its action space more uniformly rather than collapsing to a few dominant actions.

\section{Experimental Setup}
We train PPO agents using Stable-Baselines3 with the built-in CNN policy. Key hyperparameters are: learning rate $3*10^{-4}$, $gamma = 0.999$, $n\_steps = 2048$, $batch\_ size = 128$, and 10 epochs per update. Four parallel environments are run using DummyVecEnv, with one instance optionally rendered for video inspection.

For each sequence and configuration, we ran multiple training sessions for several hundred thousand timesteps on a single CPU \& GPU. Metrics are logged to TensorBoard and additionally extracted into analysis scripts that produced figures.

\section{Results}
\subsection{Effect of Anti-Spam on Action Distribution}
Before any spam controls, action counts showed a strong collapse onto A and no-op: roughly 32.1\% A presses and 28.4\% no-ops, with only 27.2\% of actions being actual movement. After applying graduated penalties and removing Start/Select, movement actions rose to 68.2\% of all actions, A/B combined fell to 24.3\%, and no-op dropped to 7.5\%.

Shannon entropy (equation \ref{eq:entropy}) of the action distribution increased from 1.21 bits (out of a 3.17 bit maximum for nine actions) to 1.82 bits (out of 2.81 for seven actions), corresponding to about an $\approx 50\%$ relative improvement in exploration efficiency. These results are visualized in Figure \ref{fig:anti-spam}, and they align qualitatively with the goal of promoting movement over idling or spamming. 

\subsection{Exploration with Per-Map Visited Mask}
We ran an ablation study comparing two variants of the environment for the exploration sequence (2nd sequence):
\begin{itemize}
    \item \textbf{No mask}: Observations are grayscale frames only.
    \item \textbf{Visited mask}: Observations include the additional per-map visited channel.
\end{itemize}
 We conducted an A/B comparision over 300k timesteps for each variant. Certain metrics like $unique\_positions$ (Count of distinct (x, y) coordinates visited), $revisit\_ratio$ (total steps / unique positions), and $exploration\_ratio$ (Percentage of reachable tiles visited in the map) were tracked per episode.

\begin{figure}[h]
    \centering
    \includegraphics[width=\linewidth]{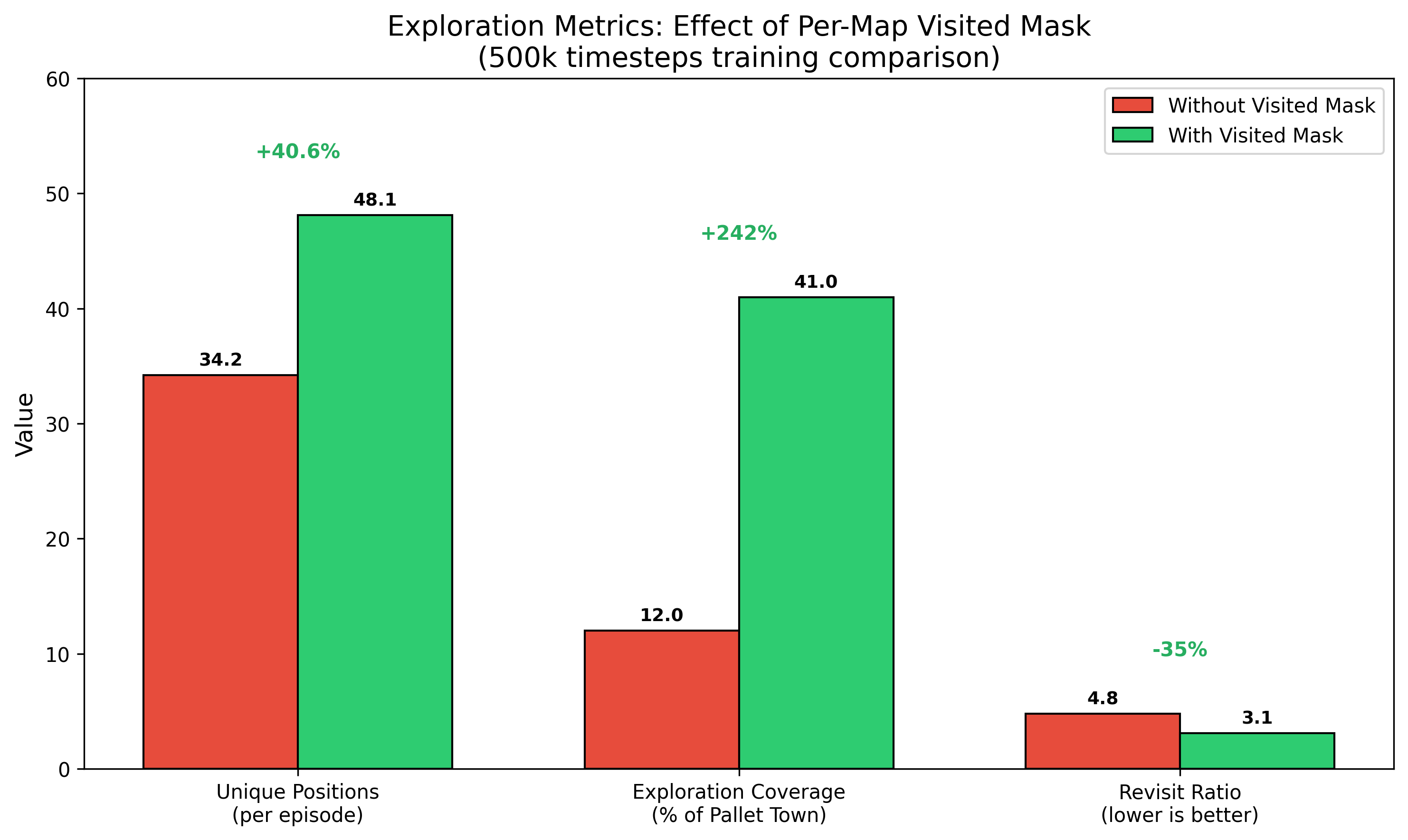}
    \caption{Exploration Metrics Comparison to see the effect of per map visited mask}
    \label{fig:exploration_metrics}
\end{figure}

 The results turned out to be the following, as highlighted by Figure \ref{fig:exploration_metrics}:
 \begin{itemize}
     \item average unique positions per episode increasing from 34.2 to 48.1 (+40.6\%).
     \item exploration coverage in Pallet Town rising from 12\% of tiles to 41\%.
     \item revisit ratio decreasing from 4.8 to 3.1, which means fewer redundant revisits.
 \end{itemize}

These differences indicate that the CNN policy effectively learned to use the mask as a memory of explored areas.

\subsection{Reduction of Loop Episodes}
Using the loop detection criteria described earlier in section \ref{section:loop}, we classified episodes as “loop episodes” if a single tile was visited more than a threshold number of times or if action patterns repeated too often. On 1,000-episode samples for each condition (as shown in Figure \ref{fig:anti-loop}):
\begin{itemize}
    \item \textbf{Baseline (no anti-loop)}: 41.2\% loop episodes, 58.8\% normal episodes.
    \item \textbf{With anti-loop system}: 4.7\% loop episodes, 95.3\% normal episodes.
\end{itemize}

This demonstrates that the multi-layer anti-loop mechanism almost completely eliminates pathological looping behaviors while leaving enough freedom for normal exploration.

\subsection{Task-Level Performance}
The current project focuses on environment design and behavior shaping rather than absolute performance, but preliminary sequence-level results are promising:

\begin{itemize}
    \item \textbf{House exit sequence}: after 150k timesteps the agent exits the house in roughly 65\% of episodes.
    \item \textbf{Exploration sequence}: the agent reaches tall grass to trigger an event, in about 60\% of episodes by 500k timesteps.
    \item \textbf{Battle sequence}: win rates against the rival hover around 50\% after 500k timesteps with a fixed reward structure.
\end{itemize}

These numbers show clear learning compared to random behavior, although there is considerable headroom before approaching human consistency.

There are some minor downsides as well. While running the testing scenarios, a manual human trigger might be required in text dialog sequences. We believe that this can be overlooked, given the current context and intent of this project.

\section{Discussion}
PokeRL demonstrates that environment-side engineering can dramatically improve the stability and interpretability of RL training in complex games. Several design choices emerged as particularly important:

\begin{itemize}
    \item \textbf{Mild penalties instead of harsh ones}: Large negative rewards for spam or loops initially collapsed training. Scaling them down by an order of magnitude preserved the incentive structure without drowning out positive signals.
    \item \textbf{Explicit modeling of game quirks}: The double-press movement mechanic is a small but essential detail. Without encoding it, the agent’s behavior looks superficially active but achieves no positional change.
    \item \textbf{Spatial memory as an observation channel}: Providing a visited mask is a simple way to give the agent memory without recurrent networks. The empirical gains in exploration metrics justify this additional channel.
    \item \textbf{Curriculum structure}: Decomposing training into house exit, exploration, and battle sequences allowed faster iteration and clearer debugging than a monolithic full-game objective would have.
\end{itemize}

At the same time, there are limitations. The reliance on direct memory access and handcrafted reward signals means our system is not fully “end-to-end” from pixels alone. Training remains computationally heavy, time-expensive, and sensitive to hyperparameters.

\section{Future Work}
A natural next step is transitioning from independently trained sequences to a unified environment where the agent must coordinate house exit, overworld navigation, and battle strategy within a single long episode. Exploration quality may be further improved by augmenting the visited mask with principled intrinsic motivation methods such as curiosity rewards or Go-Explore style state archives, which could offer more reliable long-horizon discovery than deterministic loop penalties. Another promising direction is reducing manual reward shaping through learned reward models, including inverse RL or preference-based RL guided by human feedback, which may help avoid reward hacking while preserving meaningful skill acquisition. Battle performance can also be strengthened by incorporating insights from recent language-model-guided agents that demonstrate human-parity decision making in turn-based combat. Finally, packaging the full PokeRL environment as an open benchmark with standardized tasks, metrics, and baselines would enable reproducible experimentation and foster broader community-driven progress on long-horizon JRPG-style RL challenges.

\section{Conclusion}
This paper presented PokeRL, a loop-aware reinforcement learning environment experiment for early-game Pokémon Red. By explicitly addressing infinite loops, reward sparsity, button spam, double-press movement, and memoryless exploration, we created a system in which PPO agents learn meaningful behaviors within selected objectives: exiting the house, exploring Pallet Town, and winning the first battle.

The main lesson is that successfully applying RL to rich, real games often requires as much careful system design as algorithmic innovation. Our anti-loop modules, hierarchical rewards, and visited mask are pragmatic tools that can be adapted to other long-horizon environments where agents would otherwise get lost, stuck, or bored. We hope this work contributes a concrete stepping stone on the path from toy benchmarks to agents that can robustly handle the full complexity of games like Pokémon Red.

% \section*{Acknowledgment}

% This project was initiated as a final course project for grad-level Deep Reinforcement Learning course at Texas A\&M University. I thank Prof. Dr. Sharon Guni, the instructor of this course for giving us this opportunity to explore an idea I've always wanted to tinker with. The environment design draws inspiration from the Github blog of Pokemon RL framework by David Rubinstein. All strategies for loop prevention, spam detection, exploration incentives like masking, and game mechanics handling were developed from scratch for this implementation.

\bibliographystyle{plain}
\bibliography{ref}

\end{document}